# Diagnosis of Multiple Faults: A Sensitivity Analysis

**David Heckerman**
Microsoft Research Center and
Department of Computer Science, UCLA
One Microsoft Way, 9S/1024
Redmond, WA 98052-6399
<heckerma@microsoft.com>

**Michael Shwe**
Institute for Decision Systems Research
350 Cambridge Avenue, Suite 380
Palo Alto, CA 94306
<shwe@kic.com>

## Abstract

We compare the diagnostic accuracy of three diagnostic inference models: the simple Bayes model, the multimembership Bayes model, which is isomorphic to the parallel combination function in the certainty-factor model, and a model that incorporates the noisy OR-gate interaction. The comparison is done on 20 clinicopathological conference (CPC) cases from the *American Journal of Medicine*—challenging cases describing actual patients often with multiple disorders. We find that the distributions produced by the noisy OR model agree most closely with the gold-standard diagnoses, although substantial differences exist between the distributions and the diagnoses. In addition, we find that the multimembership Bayes model tends to significantly overestimate the posterior probabilities of diseases, whereas the simple Bayes model tends to significantly underestimate the posterior probabilities. Our results suggest that additional work to refine the noisy OR model for internal medicine will be worthwhile.

## 1 INTRODUCTION

The development of practical models and inference algorithms for diagnosing multiple faults using probabilistic methods has been a long-standing challenge to researchers (Shortliffe and Buchanan, 1975; Miller et al., 1976; Reggia, 1983). An early model used for probabilistic diagnosis was the *simple Bayes model* (Ledley and Lusted, 1959). The model facilitated tractable representation and inference, by making strong assumptions about the domain. In particular, the model consists of the assumptions that diseases are mutually exclusive and exhaustive, and that findings are conditionally independent, given the presence of any disease.

In the early 1980s, Ben-Bassat developed a probabilistic model, called the *multimembership Bayes model* that relaxed the single-fault assumption (Ben-Bassat, 1980). The model includes the assumptions that diseases are marginally independent, and that findings are conditionally independent, given the presence or the absence of any disease. The model is isomorphic to the parallel combination function in MYCIN (Heckerman, 1985), an expert system for the diagnosis of bacterial infection and meningitis (Shortliffe, 1974), as well as the scoring scheme for Quick Medical Reference (QMR) (Heckerman and Miller, 1986), an expert system for internal-medicine diagnosis (Miller et al., 1986).

Several years ago, researchers developed an alternative model of multiple-fault diagnosis, in which diseases are marginally independent, findings are conditionally independent given that each disease is assigned the value absent or present, and faults interact with a common finding via a noisy OR-gate (Habbema and Hilden, 1981; Heckerman, 1989; Henrion, 1990). This model, which we will call the *noisy OR model*, offers an improvement—at least in theory—over the multimembership Bayes and simple Bayes models. Researchers have successfully used the model to translate the large QMR knowledge base (600 diseases, 4000 findings, 40,000 disease–finding interactions) to a probabilistic framework, creating a normative expert system called QMR-DT (Shwe et al., 1991; Middleton et al., 1991).

In this paper, we compare the diagnostic accuracy of these three inference models in the domain of internal medicine. In particular, we evaluate the noisy OR model for QMR-DT as well as the multimembership Bayes and simple Bayes models, also derived from the QMR-DT knowledge base. The comparison is interesting for two reasons. First, it involves a extremely large, real-world domain. Second, all three models incorporate the same probability assessments, but different assumptions of conditional independence. Thus, we can view this evaluation as a sensitivity analysis for the domain of internal medicine that determines the sensitivity of diagnostic accuracy to the model assumptions.

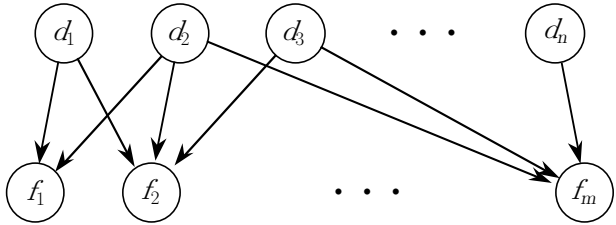

Figure 1: A belief-network encoding many of the assumptions of the noisy OR model for multiple-fault diagnosis. Diseases are marginally independent. Findings are independent, given a disease instance.

## 2 THE MODELS

In each of the models we discuss, there are $n$ diseases that may be present or absent in a patient, and $m$ findings that may be observed to be present or absent, or may not be observed at all. The problem of interest is to compute the *posterior probability* of each disease given a set of positive and negative findings.

### 2.1 NOISY OR MODEL

Several of the assumptions of the noisy OR model are shown in belief network of Figure 1. The nodes in the upper and lower layer of the network represent the diseases and findings, respectively. As indicated by the network, diseases are marginally independent, and findings are conditionally independent, given any instance of the set of diseases.[1]

Also, in this model, multiple diseases interact with a common finding via a noisy OR gate. This interaction is a special case of *causal independence* (Heckerman, 1993). Causal independence with respect to a set of diseases $d_1, \ldots, d_n$ and a single finding $f$ is represented by the belief network in Figure 2. As in Figure 1, the nodes in the upper layer of the network represent the diseases. The nodes $f_{t_0}, \ldots, f_{t_n}$ represent a temporal sequence of the findings $f$. In particular, node $f_{t_0}$ represents the finding before the patient has contracted any disease. The node $f_{t_1}$ represents the finding after the patient has (possibly) contracted disease $d_1$, but no other disease. The node $f_{t_2}$ represents the finding after the patient has (possibly) contracted diseases $d_1$ and $d_2$, but no other diseases, and so on. The node $f_{t_n}$ represents the finding after the patient has (possibly) contracted any disease; that is, $f_{t_n}$ represents the finding when it is observed. Absence of arcs in the network encode causal independence: Given $f_{t_{j-1}}$ and $d_j$, finding $f_{t_j}$ is independent of diseases $d_1, \ldots, d_{j-1}$, and findings $f_{t_k}, k = 0, \ldots, j-2$. In addition to the assumption of causal independence, the OR-gate model includes the requirements that (1) the finding and the diseases are binary, and (2) once a finding is present, it remains present regardless of other diseases that the patient might contract (hence the name, noisy OR).

[1] An instance of a set of diseases is an assignment of present or absent to each disease in that set.

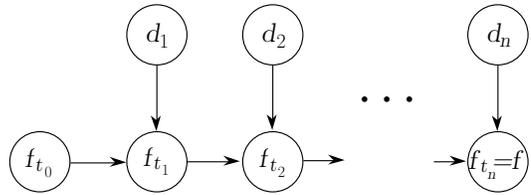

Figure 2: The noisy OR interaction between a finding $f$ and its common causes $d_1, \ldots, d_n$. Each variable in the belief network is binary. The variable $f_{t_n}$ corresponds to the observable finding $f$.

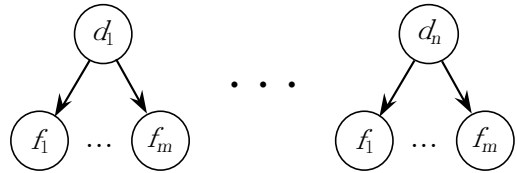

Figure 3: The multimembership Bayes model for multiple-disease diagnosis. Each disease is updated in isolation of all other diseases.

Let $d+$ and $d-$ denote the presence and absence of disease $d$, respectively. Similarly, let $f+$ and $f-$ denote the presence and absence of finding $f$, respectively. The independent probabilities in the model are the *prior probabilities* of disease, $p(d+)$, the *causal probabilities*

$$p(f_{jt_i}+|f_{jt_{i-1}}-, d_i+) \equiv q_{ij} \qquad (1)$$

and the *leak probabilities*

$$p(f_{jt_0}+) \equiv q_{0j} \qquad (2)$$

As we will see, the other models incorporate these same probabilities. In Section 3, we describe a tractable inference algorithm for this model.

### 2.2 MULTIMEMBERSHIP BAYES MODEL

In the multimembership Bayes model, we assume that diseases are marginally independent, and that all findings are independent, given that the disease is either present or absent. Furthermore, as depicted in Figure 3, we maintain a separate model for each disease. In so doing, we incorrectly ignore dependencies among findings induced by the presence of other diseases.

The probabilities required by this model are the prior probabilities of disease (the same as those in the noisy OR model), and the probabilities $p(f+|d+)$ and $p(f+|d-)$, for all findings $f$ and diseases $d$. These probabilities are computed from the noisy OR model. In particular, we have

$$p(f_j-|d_i+) = \frac{p(f_j-, d_i+)}{p(d_i+)} \qquad (3)$$

and
$$p(f_j-, d_i+) = \sum_{d_i+ \in D} p(f_j-, D) \quad (4)$$

where $D$ denotes an instance of the set of diseases in the domain. Because diseases are marginally independent in the noisy OR model, we obtain

$$p(f_j-, D) = p(f_j-|D)\, p(D) = p(f_j-|D) \prod_{k=1}^{n} p(d'_k) \quad (5)$$

where $d'_k$ denotes the instance of $d_k$ in $D$. From the noisy OR model, we know that for $f_j$ to be absent given $D$, all the diseases present in $D$ must have failed to cause $f_j$ to be present. Thus, we have

$$p(f_j-|D) = (1 - q_{0j}) \prod_{d_k+ \in D} (1 - q_{kj}) \quad (6)$$

From Equations 4 through 6, we obtain

$$p(f_j-, d_i+) = (1 - q_{0j})(1 - q_{ij})\, p(d_i+) \cdot \quad (7)$$
$$\prod_{k \neq i} [(1 - q_{kj})\, p(d_k+) + p(d_k-)]$$

Combining Equations 3 and 7, we get

$$p(f_j-|d_i+) = (1 - q_{ij})\alpha$$

where

$$\alpha = (1 - q_{0j}) \prod_{k \neq i} [(1 - q_{kj})\, p(d_k+) + p(d_k-)]$$

Therefore,

$$p(f_j+|d_i+) = 1 - (1 - q_{ij})\alpha$$

Using similar algebraic manipulations, we obtain

$$p(f_j+|d_i-) = 1 - \alpha$$

The inference algorithm for the model is straightforward. To compute the posterior probability of a particular disease, we use the odds–likelihood formulation of Bayes' theorem:

$$O(d_j+|f'_1, \ldots, f'_m) = O(d_j+) \prod_{i=1}^{n} \lambda_{ij}$$

where $O(d_j+)$ is the *prior odds* of disease $d_j$, $O(d_j+|f'_1, \ldots, f'_m)$ is the *posterior odds* of disease $d_j$ given instances $f'_1, \ldots, f'_n$, and $\lambda_{ij} = \frac{p(f'_j|d_i+)}{p(f'_j|d_i-)}$ is the *likelihood ratio* for instance $f'_j$ and disease $d_i+$.

### 2.3 SIMPLE BAYES MODEL

As mentioned, in the simple Bayes model, diseases are mutually exclusive and exhaustive, and all findings are conditionally independent given a disease. The model requires prior probabilities, which we compute by renormalizing the prior probabilities in the noisy OR model. That is,

$$p_{\mathrm{sB}}(d+) = \frac{p_{\mathrm{nO}}(d+)}{\sum_{i=1}^{n} p_{\mathrm{nO}}(d_i+)}$$

Table 1: Case information. $|F+|$ and $|F-|$ denote the number of positive and negative findings, respectively. $|D|$ denotes the number of disorders in the gold-standard diagnosis.

| Case | Source | $|F+|$ | $|F-|$ | $|D|$ |
|------|--------|--------|--------|-------|
| 1  | *AJM*:59, p241  | 51 | 8  | 4 |
| 2  | *AJM*:60, p397  | 37 | 23 | 1 |
| 3  | *AJM*:62, p616  | 27 | 13 | 2 |
| 4  | *AJM*:62, p743  | 37 | 13 | 4 |
| 5  | *AJM*:63, p273  | 41 | 18 | 3 |
| 6  | *AJM*:63, p789  | 31 | 9  | 1 |
| 7  | *AJM*:64, p651  | 23 | 10 | 2 |
| 8  | *AJM*:65, p315  | 41 | 5  | 1 |
| 9  | *AJM*:65, p63   | 32 | 16 | 4 |
| 10 | *AJM*:66, p1015 | 35 | 8  | 4 |
| 11 | *AJM*:67, p665  | 35 | 11 | 1 |
| 12 | *AJM*:68, p141  | 26 | 1  | 4 |
| 13 | *AJM*:69, p127  | 51 | 2  | 5 |
| 14 | *AJM*:69, p309  | 34 | 17 | 1 |
| 15 | *AJM*:69, p595  | 33 | 6  | 3 |
| 16 | *AJM*:69, p775  | 47 | 8  | 2 |
| 17 | *AJM*:68, p267  | 33 | 12 | 1 |
| 18 | *AJM*:68, p595  | 19 | 14 | 1 |
| 19 | *AJM*:68, p757  | 29 | 24 | 1 |
| 20 | *AJM*:68, p932  | 18 | 14 | 1 |

where $p_{\mathrm{sB}}(d+)$ is the prior probability of disease $d$ in the simple Bayes model, and $p_{\mathrm{nO}}(d+)$ is the prior probability of disease $d$ in the noisy OR model. Also, the model requires the conditional probabilities $p(f+|\text{only } d \text{ present})$, which we compute from the noisy OR model:

$$p(f_j+|\text{only } d_i \text{ present}) = 1 - (1 - q_{ij})(1 - q_{0j})$$

where $q_{ij}$ and $q_{0j}$ are defined by Equations 1 and 2, respectively.

## 3 EXPERIMENTAL DESIGN

In our evaluation, we used 20 diagnostic cases abstracted from published clinicopathological conference (CPC) cases from the *American Journal of Medicine*. CPC cases are challenging cases describing actual patients often with multiple disorders. In the 20 cases, the number of disorders in the *gold-standard diagnosis*—established by pathological investigation at autopsy—ranges from one to four. Each of these cases was abstracted by the QMR group for testing of the QMR system. We selected the first 20 cases from a set of 48 cases that we received from the QMR group. We have used these cases in previous evaluations of inference algorithms. Additional information about the test cases appears in Table 1.

We know of no tractable algorithm that can compute the exact posterior probabilities of disease using the noisy OR model for CPC cases.[2] Consequently, we used the sampling algorithm S to compute the posterior distributions (Shwe and Cooper, 1991). The algorithm uses likelihood weighting (Fung and Chang, 1989; Shachter and Peot, 1989) in combination with importance sampling (Shachter and Peot, 1989) and Markov-blanket scoring (Pearl, 1987). Each case converged within 3 hours, running on a Macintosh Quadra 950.[3] The number of samples for each case ranged from 70,000 to 100,000. Inference using the multimembership Bayes and simple Bayes models required less than 1 second per case.

## 4 RESULTS AND DISCUSSION

The results for cases 1 through 7, 8 through 14, and 15 through 20 are shown in Figures 4, 5, and 6, respectively. In each graph, the heights of the three bars associated with value $i$ on the $x$ axis correspond to $p(d_i + |\text{findings})$ for the three models, where $d_i$ is the $i$th most likely disease in the noisy OR model. The posterior probabilities for the noisy OR, multimembership Bayes, and simple Bayes models correspond to the black, white, and dotted bars, respectively. The asterisks in the figures indicate the gold-standard diagnoses. The gold-standard diagnoses for cases 2, 6, and 18 were the 116th, 212th, and 74th most likely diseases in the noisy OR model; thus, they do not appear in the figures.

The results indicate that there are substantial differences among the gold-standard diagnoses and the posterior probability distributions for the three models. Overall, the distributions produced by the noisy OR model are most in agreement with the gold-standard diagnoses. In some single-fault cases, however, the distributions produced by the simple Bayes model agree more closely with the gold-standard diagnoses (see cases 8 and 17). This result is not surprising, because the assumption that only one disease is present is built into the simple Bayes model.

The substantial differences between the OR-model distributions and the gold-standard diagnoses may be due, in part, to the fact that the gold-standard diagnoses represent outcomes and not necessarily the best posterior distributions given the evidence provided to the inference models. Nonetheless, this study provides good evidence that additional work to refine the noisy OR model for internal medicine will be worthwhile.

Two additional patterns emerge from the results: the multimembership Bayes model tends to overestimate the probability of diseases, whereas the simple Bayes model tends to underestimate the probability of diseases. For example, in case 1, there are five diseases that have posterior probabilities in the noisy OR model greater than 0.5. In contrast, 43 of the top 50 diseases have posterior probabilities in the multimembership Bayes model greater than 0.5; there are no such diseases in the simple Bayes model.

These patterns are not surprising. Because the posterior probabilities of disease must sum to 1 in the simple Bayes model, few diseases can have substantial probabilities. Thus, for cases where multiple diseases are present, the simple Bayes model will underestimate the probabilities of those diseases. The multimembership Bayes model provides no mechanism for diseases that share common findings to compete for the explanation of those findings. Consequently, the model tends to overestimate the probabilities of disease. The noisy OR model lies between the two approaches: diseases can partially, but not completely, exclude one another. Although these patterns can be predicted qualitatively, the degree of the effect in this real-world example is surprising to these authors.

## 5 Acknowledgment

This work has been supported in part by the National Science Foundation under Grant IRI-9120330.

---

[2]The Quickscore algorithm (Heckerman, 1989) is efficient for cases that contain 15 or fewer findings observed to be present, but the CPC cases contain many more such findings.

[3]Cooper and Shwe (1991) developed criteria to test for convergence. Each case in this study met those criteria.

Figure 4: The posertior probabilities of disease as a function of rank ($y$ axis) for cases 1 through 7. The noisy OR, multimembership Bayes, and simple Bayes models are depicted by black, white, and dotted bars, respectively. The asterisks indicate the gold-standard diagnoses.

Figure 5: The posertior probabilities of disease as a function of rank ($y$ axis) for cases 8 through 14. The noisy OR, multimembership Bayes, and simple Bayes models are depicted by black, white, and dotted bars, respectively. The asterisks indicate the gold-standard diagnoses.

Figure 6: The posertior probabilities of disease as a function of rank ($y$ axis) for cases 16 through 20. The noisy OR, multimembership Bayes, and simple Bayes models are depicted by black, white, and dotted bars, respectively. The asterisks indicate the gold-standard diagnoses.